\documentclass{article} 
\usepackage{iclr2027_conference,times}

\usepackage{amsmath,amsfonts,bm}

\def\eqref#1{equation~\ref{#1}}

\def\1{\bm{1}}

\DeclareMathAlphabet{\mathsfit}{\encodingdefault}{\sfdefault}{m}{sl}
\SetMathAlphabet{\mathsfit}{bold}{\encodingdefault}{\sfdefault}{bx}{n}

\usepackage{hyperref}
\usepackage{url}
\usepackage{graphicx}
\usepackage{booktabs} 
\usepackage{xcolor}
\usepackage{tabularx}
\usepackage{pifont}
\usepackage{amsmath}
\usepackage{amssymb}
\usepackage{multirow}
\usepackage{enumitem}
\usepackage{array}
\usepackage{algorithm}
\usepackage{algpseudocode}
\usepackage{colortbl}
\newcommand{\grayline}{%
  \arrayrulecolor{black!20}%
  \hline
  \arrayrulecolor{black}%
}

\newcommand{\cmark}{\textcolor{green!60!black}{\ding{51}}}
\newcommand{\xmark}{\textcolor{red}{\ding{55}}}
\definecolor{armygreen}{HTML}{6F7A3A}
\definecolor{darkyellow}{RGB}{200, 160, 0}
\definecolor{darkgreen}{RGB}{0, 150, 0}
\definecolor{darkblue}{RGB}{0, 90, 160}

\title{
    \includegraphics[height=1.2em]{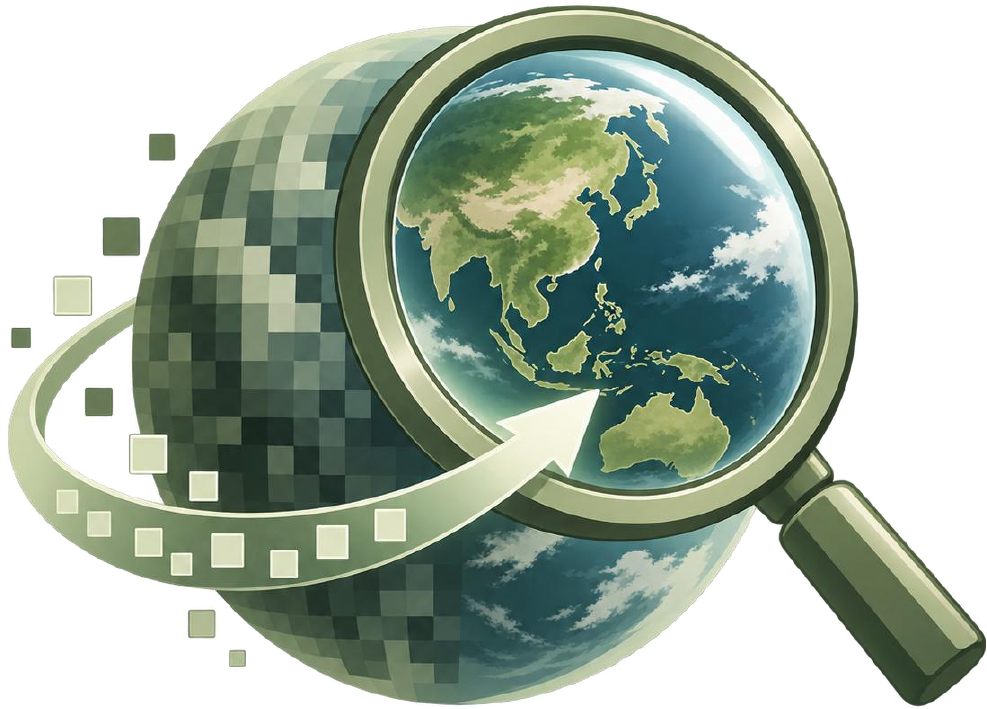} 
    \textcolor{armygreen}{RS-OPSD}: Reliable Privileged \textcolor{armygreen}{O}n-\textcolor{armygreen}{P}olicy \textcolor{armygreen}{S}elf-\textcolor{armygreen}{D}istillation for Ultra-High-Resolution \textcolor{armygreen}{R}emote \textcolor{armygreen}{S}ensing VQA
}

\author{
Chengjie Jiang$^{1,*\dagger}$ \;
Yunqi Zhou$^{2,*}$ \;
Jiafeng Yan$^{3}$ \;
Sihang Zhao$^{4,5}$ \;
Chun Yuan$^{1,\ddagger}$ \;
Jing Li$^{4,5,\dagger\ddagger}$ \\
\\
$^{1}$Tsinghua University \quad
$^{2}$Zhejiang University \quad
$^{3}$Central University of Finance and Economics \\
$^{4}$East China Normal University \quad
$^{5}$Key Laboratory of Geographic Information Science
}

\iclrfinalcopy 
\begin{document}

\maketitle
\lhead{}   
\begingroup
\renewcommand{\thefootnote}{\fnsymbol{footnote}}
\footnotetext[1]{Equal contribution.}
\footnotetext[2]{Project leaders.}
\footnotetext[3]{Corresponding authors.}
\endgroup
\vspace{-10pt}
\begin{figure}[!h]
    \centering
    \includegraphics[width=0.85\textwidth]{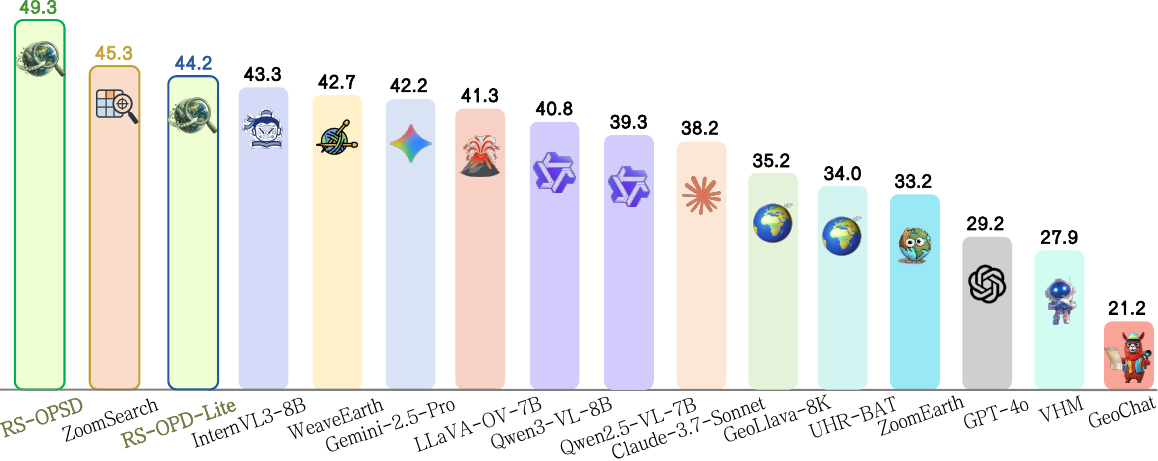}
    \caption{Average scores on Ultra-High-Resolution remote sensing VQA benchmarks, including XLRS-Bench, MME-RealWorld-RS, and LRS-VQA.}
    \label{fig:teaser}
\end{figure}

\begin{abstract}
Ultra-high-resolution (UHR) remote sensing visual question answering (VQA) requires models to resolve small visual evidence within extremely large images. Existing approaches typically rely on token pruning, visual search, or tool-augmented reasoning at inference time. We instead investigate whether the benefit of \emph{zoom-in visual privilege} can be internalized into the model. We introduce \textcolor{armygreen}{RS-OPSD}, a reliable privileged on-policy self-distillation (OPSD) framework for UHR remote sensing VQA. To provide high-quality privileged information with explicit question-relevant evidence, we construct \textbf{GeoEvidence-6K}, containing 6,750 VQA samples across seven task categories with evidence-region annotations, and develop \textbf{Human Feedback-Guided Skill Refinement (HF-SR)} for scalable annotation. To address context loss from tight crops and conflicting signals from imperfect teachers, \textcolor{armygreen}{RS-OPSD} introduces \textbf{Context-Preserving Visual Privilege (CPVP)} and \textbf{Correctness-Aligned Distillation (CAD)}. Without any additional visual search or tool calls at inference time, \textcolor{armygreen}{RS-OPSD} achieves state-of-the-art (SOTA) performance on XLRS-Bench, MME-RealWorld-RS, and LRS-VQA, outperforming previous SOTA models of comparable scale by an average of 4.0 percentage points. Moreover, our 2B variant, \textcolor{armygreen}{RS-OPD-\textit{Lite}}, surpasses most 8B-scale models while achieving the fastest measured inference speed. Our \href{https://github.com/Ronniejiang/RS-OPSD.git}{Code}, \href{https://huggingface.co/datasets/ronniejiangC/GeoEvidence-6K}{GeoEvidence-6K}, and the model weights for \href{https://huggingface.co/ronniejiangC/RS-OPSD}{\textcolor{armygreen}{RS-OPSD}} and \href{https://huggingface.co/ronniejiangC/RS-OPD-Lite}{\textcolor{armygreen}{RS-OPD-\textit{Lite}}} are publicly available.
\end{abstract}

\section{Introduction}
\label{sec:intro}
UHR remote sensing VQA requires models to reason over images containing tens of millions of pixels, where the evidence is often confined to only a tiny fraction of the scene. Although modern vision-language foundation models~\citep{qwen2.5vl,qwen3vl,llavaov,internvl3} exhibit strong semantic understanding, their effective perceptual capacity is easily overwhelmed by such extremely large visual inputs. Recent studies such as ZoomSearch~\citep{ZoomSearch} and Vision-OPD~\citep{visionopd}  suggest that many fine-grained failures are not simply caused by an inability to recognize the target itself, but by a more fundamental \emph{where-to-look} bottleneck---the model often fails to effectively attend to and exploit question-relevant visual evidence in extremely large images.

\begin{figure*}[h]
    \centering
    \includegraphics[width=0.85\textwidth]{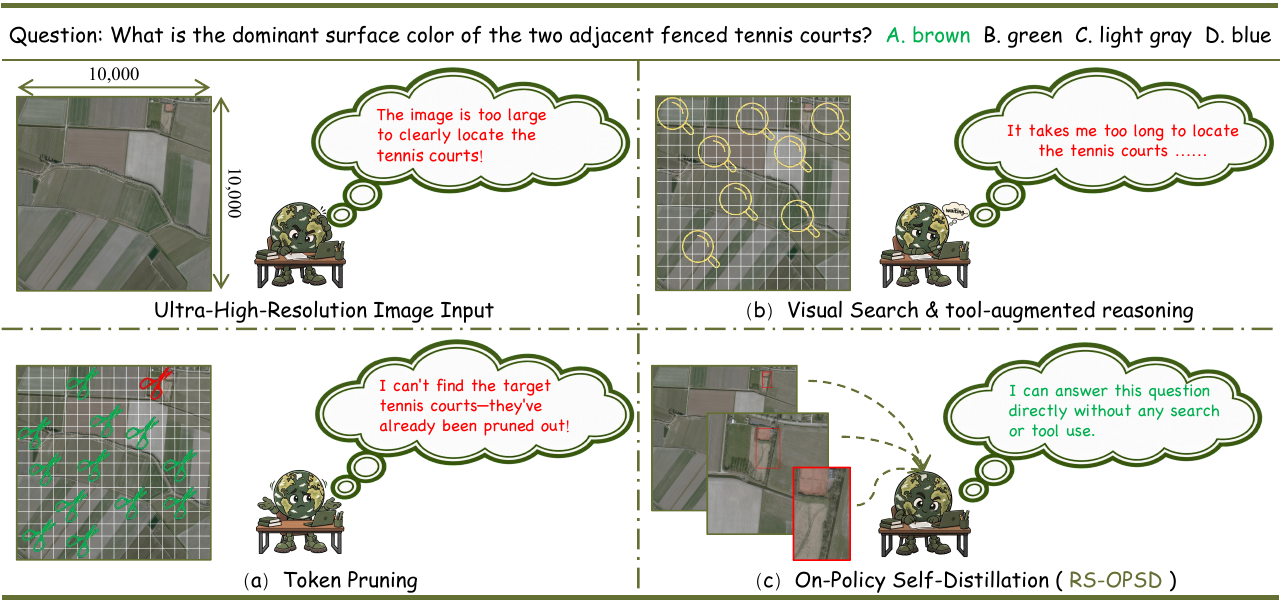}
    \caption{
    Comparison of representative paradigms for UHR remote sensing VQA.
    }
    \label{fig:intro-overview}
\end{figure*}

As illustrated in Figure~\ref{fig:intro-overview}, existing approaches mainly address this bottleneck from three directions. 
First, \emph{token-pruning} methods exploit the high redundancy of UHR imagery by retaining only a subset of visual tokens~\citep{lrsvqa,uhr-bat}. However, hard pruning is typically difficult to reverse: once evidence token is discarded, it is no longer available to subsequent reasoning. 
Second, \emph{visual search} methods search over image patches to identify relevant regions before answering~\citep{ZoomSearch,weaveearth}. While effective, such search introduces additional inference-time computation. 
Third, \emph{tool-augmented reasoning} methods allow the model to actively invoke zoom-in tools during inference, often through multiple rounds of refinement~\citep{zoomearth,geoeyes}. These methods demonstrate the benefit of exposing the model to targeted high-resolution views, but require additional tool calls and visual search. This observation motivates a different question: \emph{Can the privilege brought by visual zooming be internalized into the model during training, such that it can exploit the same fine-grained visual prior without additional search or tool use?}

A natural way to realize this idea is \emph{on-policy self-distillation} (OPSD)~\citep{opsd}. 
Instead of introducing additional visual operations at inference time, OPSD equips a teacher with privileged visual information during training and transfers this advantage to a student operating under the standard input. 
In UHR remote sensing VQA, zoomed-in evidence can naturally serve as such visual privilege: during training, the teacher is exposed to fine-grained question-relevant views, while the student learns to better exploit the corresponding visual evidence from the standard full-image input.

Realizing this training paradigm requires high-quality zoom-in evidence as visual privilege.
We therefore construct \textbf{GeoEvidence-6K}, a UHR remote sensing VQA dataset containing 6,750 samples across seven task categories, with explicit annotations for question-relevant visual evidence.
Since accurately annotating such regions requires careful inspection of extremely large images, we further develop \textbf{Human Feedback-Guided Skill Refinement (HF-SR)}, which progressively incorporates human feedback into reusable annotation skills to improve subsequent data construction.

We then conduct a pilot study using full-image inputs for the student and evidence-crop privileges for a teacher initialized from the same model. However, this straightforward formulation yields only marginal gains. We identify two major bottlenecks: first, the crop-only privilege provides only a modest advantage over the student; second, the privileged teacher itself remains imperfect, which can introduce substantial conflicting or even harmful supervision.
To address these issues, we introduce two complementary designs in \textcolor{armygreen}{RS-OPSD}. 
\textbf{Context-Preserving Visual Privilege (CPVP)} equips the teacher with global, contextual, and fine-grained evidence views, strengthening the privileged signal while preserving surrounding context. 
\textbf{Correctness-Aligned Distillation (CAD)} filters teacher supervision through sample-level reliability gating and retains only token-level update directions aligned with the correctness of the student response.

Using Qwen3-VL-8B~\citep{qwen3vl} as the base model, \textcolor{armygreen}{RS-OPSD} achieves SOTA performance on all three UHR remote sensing VQA benchmarks, outperforming the strongest competing methods by 1.6 percentage points on XLRS-Bench~\citep{xlrs}, 3.9 points on MME-RealWorld-RS~\citep{mmerealword}, and 2.0 points on LRS-VQA~\citep{lrsvqa}. 
We further instantiate the same framework with a Qwen3-VL-2B student and a privileged Qwen3-VL-8B teacher, yielding \textcolor{armygreen}{RS-OPD-\textit{Lite}}. Despite retaining only the 2B student at inference time, \textcolor{armygreen}{RS-OPD-\textit{Lite}} surpasses most 8B-scale models. Moreover, since neither variant requires additional visual search or tool calls, they both achieve low inference latency, with \textcolor{armygreen}{RS-OPD-\textit{Lite}} achieving the lowest inference latency, {12.8\% lower than the second-fastest method. 
Our main contributions are summarized as follows:

\begin{itemize}[leftmargin=*, labelsep=0.8em]
    \item We introduce \textcolor{armygreen}{RS-OPSD}, which formulates UHR remote sensing VQA as an on-policy self-distillation problem and internalizes the priviledge of zoom-in evidence into the model.
    \item To support this paradigm, we construct \textbf{GeoEvidence-6K}, a high-quality UHR remote sensing VQA dataset with explicit evidence-region annotations. We further propose \textbf{Human Feedback-Guided Skill Refinement}, which iteratively converts accumulated human feedback into reusable annotation skills for subsequent data construction.
    \item We develop two complementary components for reliable privileged distillation: \textbf{Context-Preserving Visual Privilege}, which provides the teacher with global, contextual, and fine-grained visual evidence, and \textbf{Correctness-Aligned Distillation}, which suppresses unreliable teacher supervision through sample-level reliability gating and correctness-aligned token updates.
    \item Extensive experiments show that \textcolor{armygreen}{RS-OPSD} achieves SOTA performance while maintaining efficient, search-free and tool-free inference. Moreover, \textcolor{armygreen}{RS-OPD-\textit{Lite}} retains competitive accuracy and achieves the fastest inference speed among all compared methods.
\end{itemize}
\section{Related Work}
\label{sec:relatedwork}
As shown in Figure~\ref{fig:intro-overview}, existing approaches to UHR remote sensing VQA mainly follow three paradigms: \emph{token pruning}, \emph{visual search}, and \emph{tool-augmented reasoning}.

\textbf{Token-pruning methods} exploit the redundancy in UHR remote sensing imagery by retaining question-relevant visual tokens while discarding less informative regions.
GeoLLaVA-8K~\citep{geollava8k} combines Background Token Pruning with Anchored Token Selection to suppress redundant background tokens while preserving object-centric visual tokens.
UHR-BAT~\citep{uhr-bat} further introduces query-guided multi-scale token selection and region-faithful preserve-and-merge strategies to allocate a fixed token budget toward question-relevant regions.

\textbf{Visual search methods} explicitly identify informative regions and construct compact evidence from the original UHR image before reasoning, avoiding dense processing of the entire scene.
ZoomSearch~\citep{ZoomSearch} performs hierarchical zoom search to locate question-relevant patches and reassembles the selected evidence with spatial layout preserved.
WeaveEarth~\citep{weaveearth} constructs a compact Minimal Support Evidence Set through global-aware evidence selection and integrates local evidence, spatial metadata, and relative topology for subsequent reasoning.

\textbf{Tool-augmented reasoning methods} equip VLMs with explicit visual tools, enabling active acquisition of high-resolution evidence during inference.
ZoomEarth~\citep{zoomearth} introduces an active perception paradigm that learns adaptive cropping and zooming through supervised fine-tuning and reinforcement learning, allowing the model to revisit informative regions during reasoning.
Building on this direction, GeoEyes~\citep{geoeyes} further addresses the tendency of zoom-enabled models to follow homogeneous tool-use patterns, learning more adaptive on-demand zooming and stopping behaviors through staged supervised and reinforcement learning.
\section{GeoEvidence-6K}
\label{sec:dataset-construction}
To provide the teacher with zoom-in visual privileges, we require high-quality UHR remote sensing data pairing each question with a target region. We therefore construct \textbf{GeoEvidence-6K}, a manually verified UHR remote sensing VQA dataset with explicit target-region annotations. Since annotating these regions is labor-intensive and prone to repetitive errors, we introduce \textbf{Human Feedback-Guided Skill Refinement (HF-SR)}, a human-in-the-loop annotation framework that consolidates human corrections into reusable annotation skills. Human feedback thus not only corrects individual samples but also continuously improves subsequent annotation.

\subsection{Dataset Overview}

GeoEvidence-6K is constructed from 1,050 UHR remote sensing images collected from five public data sources, as summarized in Table~\ref{tab:image-sources}. The images average $10{,}714\times10{,}714$ pixels, with ground sampling distances (GSDs) ranging from 0.08\,m to 0.50\,m, and span diverse urban, agricultural, natural, coastal, and mountainous scenes. The annotated target regions average only $1{,}095\times1{,}074$ pixels, roughly $1\%$ of the full-image area. This large scale disparity highlights the difficulty of locating and resolving question-relevant visual evidence in UHR scenes.
\vspace{-6pt}
\begin{table}[h]
\centering
\caption{Statistics of the image sources used in GeoEvidence-6K.}
\label{tab:image-sources}
\scriptsize
\setlength{\tabcolsep}{3pt}
\renewcommand{\arraystretch}{1.05}

\begin{tabularx}{\columnwidth}{
@{}
>{\raggedright\arraybackslash}X
c
c
c
c
l
@{}
}
\toprule
Source
& Nums.
& \mbox{Image Size}
& \mbox{Target Size}
& GSD
& Country \\
\midrule

MiniFrance~\citep{castillo2022minifrance}
& 110
& $10K$
& $509\!\times\!517$
& $0.50\,\mathrm{m}$
& France \\

HRSCD~\citep{daudt2019hrscd}
& 40
& $10K$
& $516\!\times\!517$
& $0.50\,\mathrm{m}$
& France \\

SWISSIMAGE~\citep{swissimage}
& 300
& $10K$
& $837\!\times\!817$
& $0.10\,\mathrm{m}$
& Switzerland \\

GeoNRW~\citep{geonrw}
& 300
& $10K$
& $997\!\times\!987$
& $0.10\,\mathrm{m}$
& Germany \\

Beeldmateriaal~\citep{beeldmateriaal}
& 300
& $12K$
& $1{,}741\!\times\!1{,}696$
& $0.08\,\mathrm{m}$
& Netherlands \\

\bottomrule
\end{tabularx}
\end{table}

The dataset covers seven multiple-choice task categories. Five are local-evidence tasks, including \texttt{counting}, \texttt{position}, \texttt{color}, \texttt{category}, and \texttt{shape}, while \texttt{land-use} and \texttt{route-planning} are global-context tasks. For each source image, we select diverse target regions and annotate questions of different task types, resulting in a total of 6,750 training samples. Representative annotations are shown in Figure~\ref{fig:dataset-pipeline}.

\begin{figure}[h]
\centering
\includegraphics[width=1.0\linewidth]{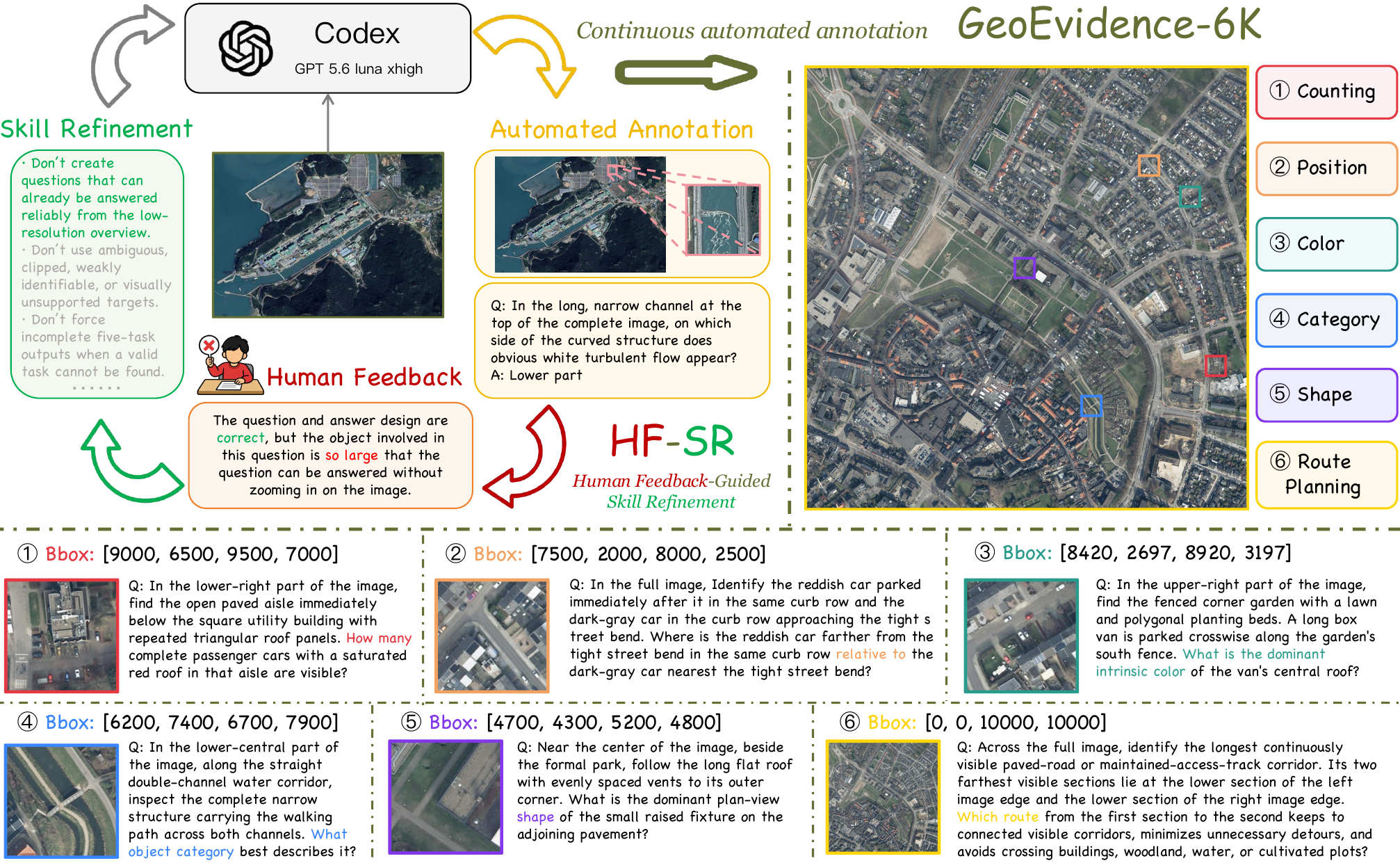}
\caption{Overview of the HF-SR pipeline, together with representative cases of GeoEvidence-6K.}
\label{fig:dataset-pipeline}
\end{figure}

\subsection{HF-SR: Human Feedback-Guided Skill Refinement}

Figure~\ref{fig:dataset-pipeline} illustrates our HF-SR pipeline. 
Let $\mathcal{A}$ denote a fixed annotation agent and $\mathcal{S}_k$ the annotation skill at refinement stage $k$. Given a UHR remote sensing image $x_i$, the agent generates $\hat{y}_i=\mathcal{A}(x_i;\mathcal{S}_k)$, which is reviewed by humans to obtain feedback $f_i=\mathcal{H}(x_i,\hat{y}_i)$. Unlike conventional human verification, HF-SR reuses accumulated feedback to update the annotation skill. Specifically, feedback collected between two checkpoints is $\mathcal{F}_k=\{f_i \mid c_k<i\leq c_{k+1}\}$, and the skill is refined as $\mathcal{S}_{k+1}=\mathcal{R}(\mathcal{S}_k,\mathcal{F}_k)$, where $\mathcal{R}$ denotes an agent-driven refinement process that summarizes recurring errors and human corrections into reusable procedural rules.

The agent follows an overview-to-detail procedure: it first inspects a downsampled full-image overview and then examines candidate regions at native resolution. For local-evidence tasks, it selects spatially diverse target regions and generates the corresponding bounding boxes, questions, and reference answers. Human reviewers jointly inspect the full image, target regions, and structured annotations, correcting issues such as ambiguous references, incomplete target regions, invalid crop geometry, or inconsistent question--answer design.

To account for the evolving maturity of the annotation skill, we schedule refinement checkpoints using cumulative Fibonacci intervals, with $c_k=\sum_{j=1}^{k}F_j$ and $F_1=F_2=1$. Frequent updates are used during the cold-start stage, when new failure patterns emerge rapidly, while progressively longer intervals are adopted as the skill stabilizes to amortize refinement cost. 
\section{Preliminary}
\label{sec:preliminary}

\textbf{On-Policy Distillation.}
On-policy distillation (OPD) transfers knowledge from a teacher policy
$\pi_T$ to a student policy $\pi_\theta$ on trajectories sampled from
the student itself. Given an input $x$, the student first generates
$\mathbf{y}=(y_1,\ldots,y_T)\sim\pi_\theta(\cdot\mid x)$.
At each student-visited prefix $\mathbf{y}_{<t}$, the student and teacher
produce next-token distributions:
\begin{equation}
p_t(\cdot)=\pi_\theta(\cdot\mid x,\mathbf{y}_{<t}),
\qquad
q_t(\cdot)=\pi_T(\cdot\mid x,\mathbf{y}_{<t}).
\end{equation}
OPD then minimizes the reverse KL divergence along the student trajectory:
\begin{equation}
\mathcal{L}_{\mathrm{OPD}}
=
\mathbb{E}_{\mathbf{y}\sim\pi_\theta(\cdot\mid x)}
\left[
\sum_{t=1}^{T}
D_{\mathrm{KL}}
\left(
p_t \,\|\, q_t
\right)
\right].
\label{eq:opd}
\end{equation}
Unlike off-policy distillation, the teacher provides supervision on states
actually visited by the current student policy, reducing train--inference
distribution mismatch.

\textbf{On-Policy Self-Distillation.}
On-policy self-distillation (OPSD) removes the need for a separate
stronger teacher by assigning different contextual roles to the same
underlying model~\citep{opsd}. The student observes the standard input
$x$, whereas the teacher is additionally conditioned on privileged
information $z$ that is available only during training:
\begin{equation}
p_t^{S}
=
\pi_\theta(\cdot\mid x,\mathbf{y}_{<t}),
\qquad
p_t^{T}
=
\pi_{\theta_T}(\cdot\mid x,z,\mathbf{y}_{<t}).
\end{equation}
The privileged teacher evaluates the same student-generated trajectory,
and the student is optimized toward its distribution:
\begin{equation}
\mathcal{L}_{\mathrm{OPSD}}
=
\mathbb{E}_{\mathbf{y}\sim\pi_\theta(\cdot\mid x)}
\left[
\sum_{t=1}^{T}
D_{\mathrm{KL}}
\left(
p_t^{S}
\,\|\, 
\operatorname{sg}\!\left[p_t^{T}\right]
\right)
\right],
\label{eq:opsd}
\end{equation}
where $\operatorname{sg}[\cdot]$ denotes stop-gradient.
In this way, OPSD transfers capabilities exposed by privileged training
information into the student while retaining the standard input at
inference time.
\section{\textcolor{armygreen}{RS-OPSD}}
\label{sec:rs-opsd}

\subsection{Motivation}
\label{subsec:pilot}

We first conduct a pilot study to examine whether zoom-in evidence can be directly exploited through OPSD for UHR remote sensing VQA. As shown in Table~\ref{tab:ablation}, LoRA fine-tuning on GeoEvidence-6K improves the baseline from 40.8 to 45.7 on average, whereas directly applying OPSD with the evidence crop as the teacher privilege achieves only 45.2. Although OPSD still improves over the base model, it fails to outperform standard LoRA fine-tuning, suggesting that crop-based privileged distillation is not yet sufficiently effective in this setting.

To understand this limitation, we further evaluate the teacher--student capability gap on GeoEvidence-6K under direct-answer evaluation. Using the same Qwen3-VL-8B model, the full-image input achieves 63.85\% accuracy, while the evidence crop reaches 69.91\%. This diagnostic reveals two bottlenecks: first, the crop privilege provides only a 6.06-point advantage, resulting in a relatively weak privileged signal; second, the privileged model itself remains imperfect, with nearly 30\% of the training samples answered incorrectly, which can introduce conflicting supervision.
These observations motivate two complementary designs in \textcolor{armygreen}{RS-OPSD}: \textbf{Context-Preserving Visual Privilege (CPVP)} enriches the teacher with additional contextual and global information to strengthen its privileged advantage, while \textbf{Correctness-Aligned Distillation (CAD)} filters unreliable teacher signals and retains only update directions aligned with verified answer correctness.

\label{sec:rs-opsd}

\begin{figure*}[h]
    \centering
        \includegraphics[width=\textwidth]{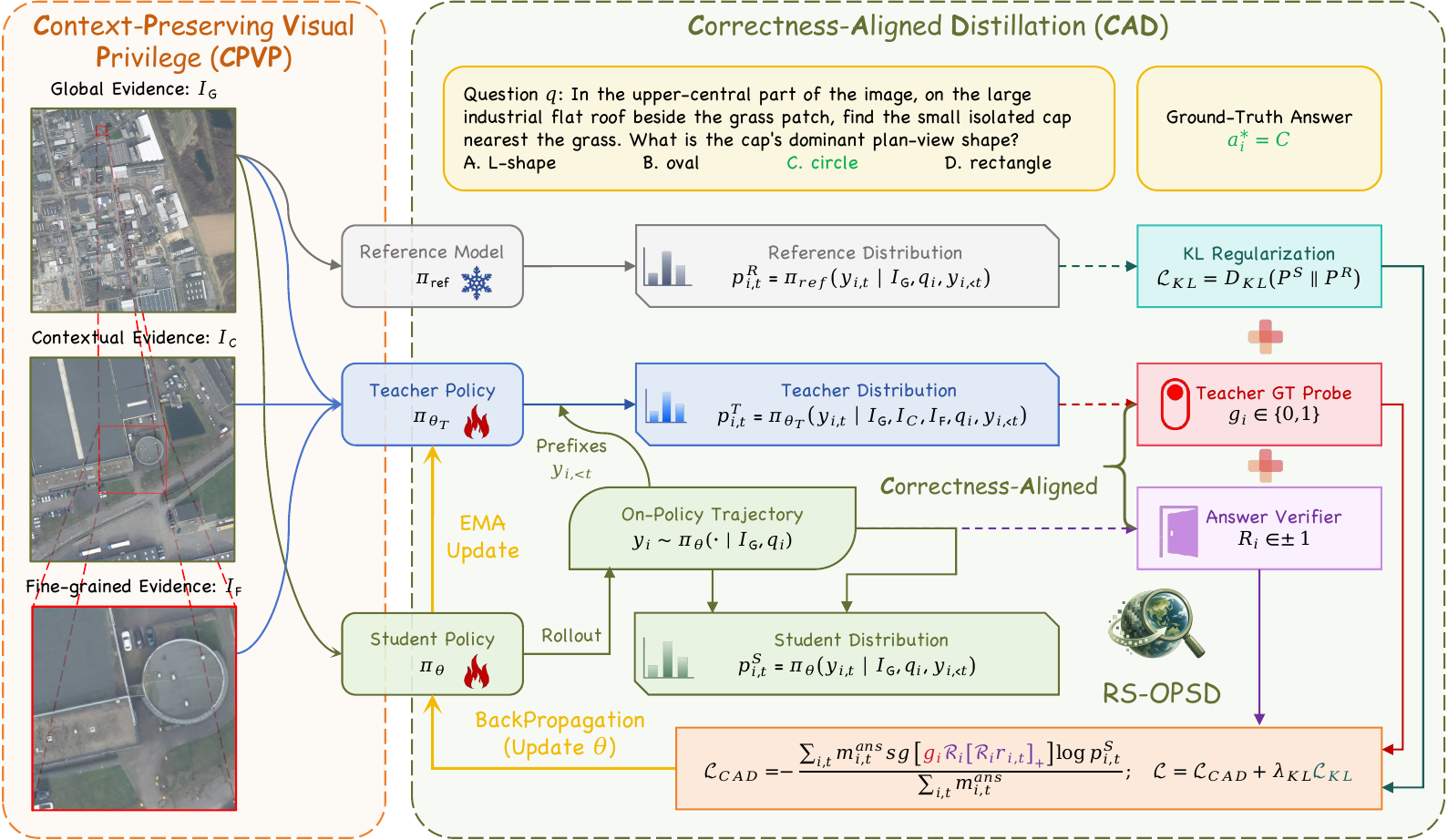}
    \caption{
    Overview of \textcolor{armygreen}{RS-OPSD}, consisting of CPVP and CAD.
    }
    \label{fig:rs-opsd}
\end{figure*}
\vspace{-6pt}

\subsection{CPVP: What Should the Teacher See?}

Building on the OPSD formulation in Section~\ref{sec:preliminary}, each training sample contains a UHR remote sensing image $I_i$, a question $q_i$, a ground-truth answer $a_i^*$, and an annotated evidence region $B_i$. In our setting, the standard input $x$ corresponds to the global evidence $I_G$, while the privileged information $z$ is constructed from the annotated evidence region and provided only to the teacher.

A tight evidence crop improves target visibility but may discard surrounding context. Conversely, the global image preserves scene-level information but weakens fine-grained targets. As shown in Figure~\ref{fig:rs-opsd}, we therefore propose \textbf{CPVP}, which provides the teacher with three complementary views: Global Evidence $I_G$, Contextual Evidence $I_C$, and Fine-grained Evidence $I_F$. Specifically, $I_F$ is the tight crop corresponding to $B_i$, while $I_C$ is obtained by keeping the same crop center and doubling its width and height to include surrounding context. The resulting privileged information is $z_i=\{I_{C,i},I_{F,i}\}$. 
The student observes the standard input $(I_{G,i},q_i)$, while the teacher is additionally conditioned on the privileged information $z_i=\{I_{C,i},I_{F,i}\}$.
CPVP thus strengthens the teacher privilege with fine-grained evidence and surrounding context while preserving the global input.

\subsection{CAD: When Should the Student Trust It?}

Although CPVP provides the teacher with richer visual evidence, privileged information does not guarantee reliable supervision. We therefore propose \textbf{CAD}, which filters privileged supervision at both the sample and token levels.
Following the on-policy formulation in Section~\ref{sec:preliminary}, the privileged teacher evaluates the same student-generated trajectory. For each sampled token $y_{i,t}$, we define the teacher--student preference as
\begin{equation}
r_{i,t}
=
\operatorname{sg}
\left[
\log p^T_{i,t}
-
\log p^S_{i,t}
\right],
\label{eq:teacher-student-gap}
\end{equation}
where $p^S_{i,t}$ and $p^T_{i,t}$ denote the probabilities assigned to the sampled token $y_{i,t}$ under the student and privileged-teacher distributions defined in Section~\ref{sec:preliminary}, respectively.


\textbf{Teacher reliability gate.}
Before using these token-level preferences, CAD first evaluates whether the privileged teacher provides reliable supervision for the current sample. Given the GT response $a_i^*$, we perform a teacher-forced prefix probe and define
\begin{equation}
g_i
=
\prod_{k}
\mathbf{1}
\left[
\tau_{i,k}^*
=
\arg\max_{u\in\mathcal{C}_{i,k}}
\pi_{\theta_T}
\left(
u
\mid
I_{G,i}, z_i, q_i, \boldsymbol{\tau}_{i,<k}^*
\right)
\right],
\label{eq:teacher-reliability}
\end{equation}
where $\boldsymbol{\tau}_i^*$ denotes the tokenized GT response and $\mathcal{C}_{i,k}$ denotes the valid continuations at position $k$. Thus, $g_i=1$ only when the teacher remains aligned with the GT response throughout the probe.

\textbf{Correctness-aligned token filtering.}
We further use the correctness of the student response to determine which teacher preferences should be retained:
\begin{equation}
R_i=
\begin{cases}
+1, & \hat{a}_i=a_i^*,\\
-1, & \hat{a}_i\neq a_i^*.
\end{cases}
\end{equation}
The resulting CAD signal is
\begin{equation}
A_{i,t}
=
g_i R_i
\left[R_i r_{i,t}\right]_+ .
\label{eq:cad-signal}
\end{equation}
When the student response is correct, CAD retains teacher preferences that reinforce the sampled tokens; when the response is incorrect, it retains preferences that suppress them. Teacher signals that are unreliable or inconsistent with verified answer correctness are discarded.

The CAD objective is
\begin{equation}
\mathcal{L}_{\mathrm{CAD}}
=
-
\frac{
\sum_{i,t}
m^{\mathrm{ans}}_{i,t}\,
\operatorname{sg}[A_{i,t}]
\log p^{S}_{i,t}
}{
\sum_{i,t}m^{\mathrm{ans}}_{i,t}
},
\label{eq:cad}
\end{equation}
where $m^{\mathrm{ans}}_{i,t}$ is a binary mask selecting valid answer tokens.

Finally, we regularize the student toward a frozen reference model that receives the same standard input $(I_{G,i}, q_i)$ as the student. The overall training objective is
\begin{equation}
\mathcal{L}
=
\mathcal{L}_{\mathrm{CAD}}
+
\lambda_{\mathrm{KL}}\mathcal{L}_{\mathrm{KL}},
\label{eq:overall-loss}
\end{equation}
where $\lambda_{\mathrm{KL}}$ controls the strength of reference regularization. We further apply gradient-norm clipping with threshold $c_{\mathrm{grad}}$ to stabilize optimization.

\section{Experiments}
\label{sec:experiments}

\subsection{Experimental Settings}

\textbf{Model Training.}
\textcolor{armygreen}{RS-OPSD} is initialized from Qwen3-VL-8B-Instruct and trained exclusively on GeoEvidence-6K. We set the KL regularization coefficient $\lambda_{\mathrm{KL}}=1\times10^{-3}$, the gradient clipping threshold $c_{\mathrm{grad}}=5$, and the global batch size to $96$. Training lasts for 150 steps. The teacher is initialized from the student and updated via exponential moving average (EMA) throughout training.
For \textcolor{armygreen}{RS-OPD-\textit{Lite}}, the student is initialized from Qwen3-VL-2B-Instruct, while Qwen3-VL-8B-Instruct serves as the privileged teacher and remains frozen during training. All other settings follow \textcolor{armygreen}{RS-OPSD}, except that \textcolor{armygreen}{RS-OPD-\textit{Lite}} is trained for 120 steps.

\textbf{Benchmarks.}
We evaluate our models on three UHR remote sensing VQA benchmarks: XLRS-Bench~\citep{xlrs}, MME-RealWorld-RS~\citep{mmerealword}, and LRS-VQA~\citep{lrsvqa}.
XLRS-Bench has an average image resolution of $8{,}500\times8{,}500$ pixels and evaluates fine-grained perception and reasoning tasks.
MME-RealWorld-RS has an average resolution of $5{,}602\times4{,}445$ pixels and contains 3,738 single-choice questions covering color, counting, and position.
LRS-VQA contains 1,657 images with an average resolution of $7{,}099\times6{,}329$ pixels and 7,333 question--answer pairs across eight categories.
We follow the official evaluation protocols of each benchmark and report the corresponding task-averaged scores.

\textbf{Baselines.}
We compare against 14 representative baselines spanning four categories.
(1) \emph{Closed-source VLMs}: GPT-4o, Claude 3.7 Sonnet, and Gemini 2.5 Pro.
(2) \emph{Open-source VLMs}: LLaVA-OV-7B, Qwen2.5-VL-7B, Qwen3-VL-8B, and InternVL3-8B.
(3) \emph{Remote sensing VLMs}: GeoChat and VHM.
(4) \emph{Specialized methods for UHR remote sensing VQA}, including token-pruning approaches such as GeoLLaVA-8K and UHR-BAT, visual-search approaches such as ZoomSearch and WeaveEarth, and tool-augmented reasoning approaches represented by ZoomEarth.

\begin{table}[h]
\centering
\caption{
Quantitative comparison results.
\textcolor{darkgreen}{Green}, \textcolor{darkyellow}{yellow}, and \textcolor{darkblue}{blue} denote the first-, second-, and third-best results.
Speed is the average inference time across the three benchmarks (s/sample).
}
\label{tab:main_results}

\setlength{\tabcolsep}{5.5pt}
\renewcommand{\arraystretch}{1.12}

\resizebox{\textwidth}{!}{
\begin{tabular}{l l l l c c c c c}
\toprule
\textbf{Method}
& \textbf{Pub.}
& \textbf{Backbone}
& \textbf{Param.}
& \textbf{XLRS}
& \textbf{MME}
& \textbf{LRS}
& \textbf{Avg.}
& \textbf{Speed}
\\
\midrule

\multicolumn{9}{c}{\textit{Closed-Source Vision Language Models}} \\
\midrule

GPT-4o~\citep{gpt4o}
& -- & -- & --
& 32.4 & 28.2 & 27.1 & 29.2 & -\\

Claude 3.7 Sonnet~\citep{Claude}
& -- & -- & --
& 40.5 & 45.7 & 28.4 & 38.2 & -\\

Gemini 2.5 Pro~\citep{gemini}
& -- & -- & --
& 44.1 & 51.7 & \textcolor{darkblue}{\textbf{30.9}} & 42.2 & -\\

\midrule
\multicolumn{9}{c}{\textit{Open-Source Vision Language Models}} \\
\midrule

LLaVA-OV-7B~\citep{llavaov}
& TMLR'25
& LLaVA-OV
& 7B
& 43.4 & 53.3 & 27.3 & 41.3 & 2.38\\

Qwen2.5-VL-7B~\citep{qwen2.5vl}
& arXiv'25
& Qwen2.5-VL
& 7B
& 47.4 & 41.0 & 29.4 & 39.3 & 1.87\\

Qwen3-VL-8B~\citep{qwen3vl}
& arXiv'25
& Qwen3-VL
& 8B
& \textcolor{darkblue}{\textbf{50.5}} & 41.9 & 30.1 & 40.8 & 1.75\\

InternVL3-8B~\citep{internvl3}
& arXiv'25
& InternVL3
& 8B
& 45.6 & 55.0 & 29.2 & 43.3 & \textcolor{darkyellow}{\textbf{1.25}}\\

\midrule
\multicolumn{9}{c}{\textit{Remote Sensing Vision Language Models}} \\
\midrule

GeoChat~\citep{geochat}
& CVPR'24
& LLaVA-v1.5
& 7B
& 22.9 & 21.3 & 19.5 & 21.2 & 1.33\\

VHM~\citep{vhm}
& AAAI'25
& Vicuna
& 7B
& 32.8 & 24.1 & 26.7 & 27.9 & 2.64\\

\midrule
\multicolumn{9}{c}{\textit{Specialized Methods for UHR Remote Sensing VQA}} \\
\midrule

GeoLLaVA-8K~\citep{geollava8k}
& NIPS'25
& LongVA
& 7B
& \textcolor{darkyellow}{\textbf{51.5}} & 28.4 & 25.8 & 35.2 & \textcolor{darkblue}{\textbf{1.30}}\\

UHR-BAT~\citep{uhr-bat}
& ICML'26
& LongVA
& 7B
& 48.6 & 33.3 & 20.0 & 34.0 & 7.40\\

\addlinespace[2pt]
\grayline
\addlinespace[2pt]

ZoomSearch~\citep{ZoomSearch}
& arXiv'25
& LLaVA-OV
& 7B
& 48.0 & \textcolor{darkyellow}{\textbf{57.6}} & 30.2 & \textcolor{darkyellow}{\textbf{45.3}} & 43.5\\

WeaveEarth~\citep{weaveearth}
& MM'26
& Qwen3-VL
& 8B
& 49.5 & 47.2 & \textcolor{darkyellow}{\textbf{31.3}} & 42.7 & 2.94\\

\addlinespace[2pt]
\grayline
\addlinespace[2pt]

ZoomEarth~\citep{zoomearth}
& CVPR'26
& Qwen2.5-VL
& 3B
& 39.2 & 38.7 & 21.6 & 33.2 & 14.2\\

\midrule
\multicolumn{9}{c}{\textit{Ours}} \\
\midrule

\textcolor{armygreen}{RS-OPD-\textit{Lite}}
& --
& Qwen3-VL
& 2B
& 45.8 & \textcolor{darkblue}{\textbf{56.2}} & 30.5 & \textcolor{darkblue}{\textbf{44.2}} & \textcolor{darkgreen}{\textbf{1.09}}\\

\textcolor{armygreen}{RS-OPSD}
& --
& Qwen3-VL
& 8B
& \textcolor{darkgreen}{\textbf{53.1}}
& \textcolor{darkgreen}{\textbf{61.5}}
& \textcolor{darkgreen}{\textbf{33.3}}
& \textcolor{darkgreen}{\textbf{49.3}} & 1.58\\

\bottomrule
\end{tabular}
}
\end{table}

\subsection{Comparison with State-of-the-Art Methods}

\textbf{Overall performance.}
As shown in Table~\ref{tab:main_results}, \textcolor{armygreen}{RS-OPSD} achieves the best performance on all three benchmarks, reaching 53.1 on XLRS-Bench, 61.5 on MME-RealWorld-RS, and 33.3 on LRS-VQA, with an average score of 49.3, exceeding the strongest competing method by 4.0 points. Detailed category-wise results and analyses are provided in Appendix~\ref{sec:detailed}.

\textbf{Accuracy--efficiency trade-off.}
\textcolor{armygreen}{RS-OPSD} runs at 1.58 s/sample, even faster than Qwen3-VL-8B baseline, while avoiding the substantial overhead of visual search and tool calls. More notably, \textcolor{armygreen}{RS-OPD-\textit{Lite}} achieves an average score of 44.2 with only 2B parameters, outperforming most 8B-scale models. It is also the fastest method at 1.09 s/sample, with $12.8\%$ lower latency than the second-fastest method. These results demonstrate that internalizing privileged visual information through OPSD can improve UHR remote sensing VQA while preserving both accuracy and efficiency.

\subsection{Ablation Studies and Analysis}

\textbf{Core design analysis.}
Table~\ref{tab:ablation} evaluates the contribution of the core designs in \textcolor{armygreen}{RS-OPSD}. 
LoRA~\citep{lora} fine-tuning on GeoEvidence-6K improves the average score of Qwen3-VL-8B from 40.8 to 45.7, indicating the high quality of the \textbf{GeoEvidence-6K}. 
In contrast, directly applying OPSD with crop-only privilege in our pilot setting slightly reduces the average score to 45.2, confirming that naïve privileged distillation is limited by the insufficient teacher--student gap and imperfect teacher supervision.
Introducing CPVP raises the average score to 47.2, while CAD independently improves it to 47.0, showing that richer visual privilege and correctness-aligned supervision both contribute positively to mitigating these two bottlenecks.
Combining CPVP and CAD further yields the best performance of 49.3, reaching the state-of-the-art results reported in Table~\ref{tab:main_results}.

\begin{table}[!t]
\centering
\caption{Ablation study of the core designs in RS-OPSD.}
\label{tab:ablation}

\fontsize{6.8}{7.8}\selectfont
\setlength{\tabcolsep}{1.8pt}
\renewcommand{\arraystretch}{1.10}

\begin{tabularx}{\columnwidth}{
@{}
c c c
>{\centering\arraybackslash}X
c c c c
@{}
}
\toprule

\multicolumn{3}{c}{\textbf{Core Design}}
& \multirow{2}{*}{\textbf{Label}}
& \multicolumn{4}{c}{\textbf{Benchmarks}} \\

\cmidrule(lr){1-3}
\cmidrule(lr){5-8}

\textbf{OPSD}
& \textbf{CPVP}
& \textbf{CAD}
&
& \textbf{XLRS}
& \mbox{\textbf{MME-RW-RS}}
& \textbf{LRS-VQA}
& \textbf{Avg.} \\

\midrule

\multicolumn{3}{c}{Qwen3-VL-8B-Instruct}
& Base Model
& 50.5 & 41.9 & 30.1 & 40.8 \\

\midrule

\xmark & \xmark & \xmark
& LoRA Finetuning on GeoEvidence-6K
& 49.9 & 55.2 & \textcolor{darkyellow}{\textbf{32.1}} & 45.7 \\

\cmark & \xmark & \xmark
& Naïve OPSD in Subsec.~\ref{subsec:pilot}
& 50.5 & 53.8 & 31.3 & 45.2 \\

\cmark & \cmark & \xmark
& OPSD + CPVP
& \textcolor{darkyellow}{\textbf{51.3}}
& \textcolor{darkyellow}{\textbf{59.0}}
& 31.2
& \textcolor{darkyellow}{\textbf{47.2}} \\

\cmark & \xmark & \cmark
& OPSD + CAD
& \textcolor{darkblue}{\textbf{51.2}}
& \textcolor{darkblue}{\textbf{58.3}}
& \textcolor{darkblue}{\textbf{31.5}}
& \textcolor{darkblue}{\textbf{47.0}} \\

\cmark & \cmark & \cmark
& \textcolor{armygreen}{RS-OPSD}
& \textcolor{darkgreen}{\textbf{53.1}}
& \textcolor{darkgreen}{\textbf{61.5}}
& \textcolor{darkgreen}{\textbf{33.3}}
& \textcolor{darkgreen}{\textbf{49.3}} \\

\bottomrule
\end{tabularx}
\end{table}

\textbf{Hyperparameter analysis.}
Table~\ref{tab:hyperparameter} examines the effects of the KL coefficient $\lambda_{\mathrm{KL}}$ and gradient clipping threshold $c_{\mathrm{grad}}$.
KL regularization consistently improves over the setting with $\lambda_{\mathrm{KL}}=0$, with $\lambda_{\mathrm{KL}}=10^{-3}$ achieving the best average score of 49.0.
Increasing $\lambda_{\mathrm{KL}}$ to $10^{-2}$ slightly degrades performance, suggesting that overly strong regularization may constrain adaptation to the privileged teacher.
With $\lambda_{\mathrm{KL}}=10^{-3}$ fixed, $c_{\mathrm{grad}}=5$ performs best at 49.3, while $c_{\mathrm{grad}}=20$ and $100$ both decrease the average to 48.3.
We therefore use $\lambda_{\mathrm{KL}}=10^{-3}$ and $c_{\mathrm{grad}}=5$ in all main experiments.

\begin{table}[!t]
\centering
\caption{
Hyperparameter studies of KL regularization and gradient clipping.
}
\label{tab:hyperparameter}

\setlength{\tabcolsep}{5pt}
\renewcommand{\arraystretch}{1.12}

\resizebox{\textwidth}{!}{
\begin{tabular}{
cc cccc
@{\hspace{10pt}}
cc cccc
}
\toprule

\multicolumn{2}{l}{\textbf{KL Coefficient}}
& \multicolumn{4}{c}{\textbf{Benchmarks}}
&
\multicolumn{2}{l}{\textbf{Clipping Threshold}}
& \multicolumn{4}{c}{\textbf{Benchmarks}}
\\

\cmidrule(lr){1-2}
\cmidrule(lr){3-6}
\cmidrule(lr){7-8}
\cmidrule(lr){9-12}

$\boldsymbol{c_{\mathrm{grad}}}$
& $\boldsymbol{\lambda_{\mathrm{KL}}}$
& \textbf{XLRS}
& \textbf{MME-RW-RS}
& \textbf{LRS-VQA}
& \textbf{Avg.}
&
$\boldsymbol{c_{\mathrm{grad}}}$
& $\boldsymbol{\lambda_{\mathrm{KL}}}$
& \textbf{XLRS}
& \textbf{MME-RW-RS}
& \textbf{LRS-VQA}
& \textbf{Avg.}
\\

\cmidrule(r){1-6}
\cmidrule(l){7-12}

1
& 0
& 51.3
& 59.0
& 31.2
& 47.2
&
5
& $1\times10^{-3}$
& \textcolor{darkyellow}{\textbf{53.1}}
& \textcolor{darkyellow}{\textbf{61.5}}
& \textcolor{darkblue}{\textbf{33.3}}
& \textcolor{darkgreen}{\textbf{49.3}}
\\

1
& $1\times10^{-2}$
& 51.2
& \textcolor{darkgreen}{\textbf{62.0}}
& 32.8
& \textcolor{darkblue}{\textbf{48.7}}
&
20
& $1\times10^{-3}$
& 51.0
& 60.0
& \textcolor{darkgreen}{\textbf{34.0}}
& 48.3
\\

1
& $1\times10^{-3}$
& \textcolor{darkgreen}{\textbf{53.2}}
& \textcolor{darkblue}{\textbf{60.8}}
& 33.1
& \textcolor{darkyellow}{\textbf{49.0}}
&
100
& $1\times10^{-3}$
& \textcolor{darkblue}{\textbf{51.5}}
& 59.9
& \textcolor{darkyellow}{\textbf{33.4}}
& 48.3
\\

\bottomrule
\end{tabular}
}
\end{table}

\begin{figure*}[h]
    \centering
    \includegraphics[width=\textwidth]{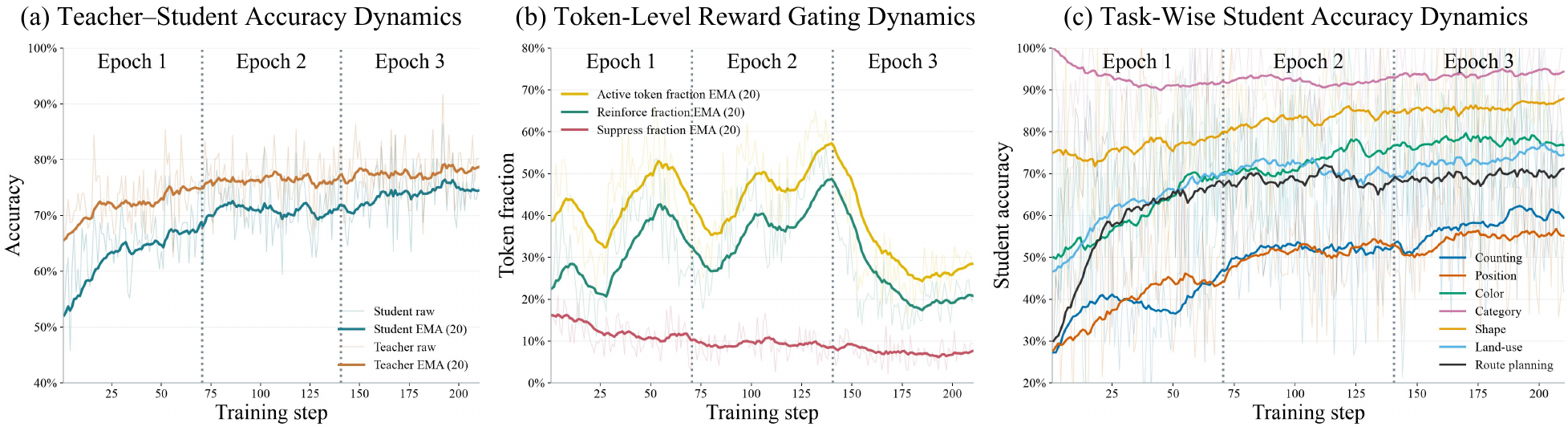}
    \caption{
    Training dynamics of \textcolor{armygreen}{RS-OPSD} across three epochs.
    }
    \label{fig:training-dynamics}
\end{figure*}

\textbf{Training dynamics analysis.}
Figure~\ref{fig:training-dynamics} further examines the training dynamics of \textcolor{armygreen}{RS-OPSD} over three epochs.
Figure~\ref{fig:training-dynamics}(a) tracks the accuracy of the student and EMA teacher. Both improve steadily throughout training, while the teacher consistently remains stronger and the teacher--student gap gradually narrows, indicating progressive transfer of privileged knowledge to the student.
Figure~\ref{fig:training-dynamics}(b) shows that suppressive token activation steadily decreases, while reinforced and overall active tokens drop sharply in the third epoch, suggesting that fewer teacher signals remain informative as the student approaches the teacher.
Finally, Figure~\ref{fig:training-dynamics}(c) reports task-wise student accuracy. Nearly all task categories exhibit consistent improvements across training.

\section{Conclusion}
\label{sec:conclusion}

In this work, we present \textcolor{armygreen}{RS-OPSD}, a reliable privileged OPSD framework for UHR remote sensing VQA. Instead of relying on additional visual search or tool use at inference time, \textcolor{armygreen}{RS-OPSD} internalizes zoom-in visual privilege into the model during training. To support this paradigm, we construct \textbf{GeoEvidence-6K} with explicit evidence-region annotations and introduce \textbf{HF-SR} for human-feedback-guided data construction. We further develop \textbf{CPVP} to provide richer contextual and fine-grained visual privilege, and \textbf{CAD} to suppress unreliable teacher guidance. Extensive experiments demonstrate that \textcolor{armygreen}{RS-OPSD} achieves SOTA performance while maintaining efficient inference, and that \textcolor{armygreen}{RS-OPD-\textit{Lite}} further offers a favorable accuracy--efficiency trade-off with a substantially smaller student model. We further discuss the limitations of the current framework and potential directions for future work in Appendix~\ref{sec:limitations}.

\bibliography{iclr2027_conference}
\bibliographystyle{iclr2027_conference}

\appendix
\section{Algorithm of \textcolor{armygreen}{RS-OPSD}}
\label{sec:algorithm}

Algorithm~\ref{alg:rs-opsd} summarizes the training procedure of
\textcolor{armygreen}{RS-OPSD}. For each sample, the student generates an
on-policy response from the Global Evidence, while the privileged teacher
evaluates the same trajectory using CPVP. CAD then filters privileged
supervision according to teacher reliability and student-answer correctness.
The student is optimized with the CAD objective and reference-model
regularization, while the teacher is updated by EMA.

\begin{algorithm}[h]
\caption{Training procedure of \textcolor{armygreen}{RS-OPSD}}
\label{alg:rs-opsd}
\begin{algorithmic}[1]

\Require Training set
$\mathcal{D}=\{(I_G,I_C,I_F,q,a^*)\}$;
student $\pi_{\theta}$;
teacher $\pi_{\theta_T}$;
frozen reference $\pi_{\mathrm{ref}}$;
KL coefficient $\lambda_{\mathrm{KL}}$;
gradient clipping threshold $c_{\mathrm{grad}}$;
EMA coefficient $\beta$

\State Initialize $\theta_T \leftarrow \theta$
\State Freeze $\pi_{\mathrm{ref}}$

\For{each training batch $\mathcal{B}\subset\mathcal{D}$}

    \For{each $(I_G,I_C,I_F,q,a^*)\in\mathcal{B}$}

        \State Construct privileged information
        $z \leftarrow \{I_C,I_F\}$

        \State Sample one student trajectory
        $\mathbf{y}\sim\pi_{\theta}(\cdot\mid I_G,q)$

        \State Construct the valid-answer mask
        $m^{\mathrm{ans}}$ from $\mathbf{y}$

        \State Compute student token probabilities
        $p_t^S \leftarrow
        \pi_{\theta}(y_t\mid I_G,q,\mathbf{y}_{<t})$

        \State Compute privileged-teacher probabilities
        $p_t^T \leftarrow
        \pi_{\theta_T}(y_t\mid I_G,z,q,\mathbf{y}_{<t})$

        \State Run the teacher-forced GT-prefix reliability probe:
        $g \leftarrow \operatorname{Probe}_T(a^*)\in\{0,1\}$

        \State Verify the student response:
        $R\leftarrow +1$ if $\hat{a}=a^*$, otherwise $R\leftarrow -1$

        \For{each valid answer token $y_t$}
            \State
            $r_t\leftarrow
            \operatorname{sg}
            [\log p_t^T-\log p_t^S]$

            \State
            $A_t\leftarrow
            g\,R\,[Rr_t]_+$
        \EndFor

    \EndFor

    \State Compute CAD loss
    \[
    \mathcal{L}_{\mathrm{CAD}}
    =
    -
    \frac{
    \sum_{i,t}
    m^{\mathrm{ans}}_{i,t}\,
    \operatorname{sg}[A_{i,t}]
    \log p^{S}_{i,t}
    }{
    \sum_{i,t}m^{\mathrm{ans}}_{i,t}
    }
    \]

    \State Compute reference-model regularization
    $\mathcal{L}_{\mathrm{KL}}$

    \State
    $\mathcal{L}
    \leftarrow
    \mathcal{L}_{\mathrm{CAD}}
    +
    \lambda_{\mathrm{KL}}\mathcal{L}_{\mathrm{KL}}$

    \State Backpropagate $\nabla_{\theta}\mathcal{L}$

    \State Clip the student gradient norm with threshold
    $c_{\mathrm{grad}}$

    \State Update student parameters $\theta$

    \State Update teacher by EMA:
    $\theta_T
    \leftarrow
    \beta\theta_T+(1-\beta)\theta$

\EndFor

\end{algorithmic}
\end{algorithm}

For \textcolor{armygreen}{RS-OPD-\textit{Lite}}, the same training
procedure is used with a Qwen3-VL-2B student and a Qwen3-VL-8B privileged
teacher. The teacher remains frozen throughout training.
\section{Dataset Quality and Integrity}
\label{sec:appendix-data-integrity}

We verified that GeoEvidence-6K has no exact image-level overlap with XLRS-Bench~\citep{xlrs}, MME-RealWorld-RS~\citep{mmerealword}, or LRS-VQA~\citep{lrsvqa}, ruling out direct train--test leakage from reused evaluation images. The annotation process involved ten domain experts with expertise in remote sensing image interpretation. All generated samples were manually reviewed, and the complete dataset underwent a final cross-validation stage across the expert pool to resolve remaining ambiguities and annotation errors. We measure annotation reliability using the Cross-Validation Acceptance Rate (CVAR), defined as the proportion of samples accepted during the final cross-validation without requiring further correction. GeoEvidence-6K achieves an overall CVAR of \textbf{97.8\%}, indicating a high level of consistency in the final annotations.

Figure~\ref{fig:dataset-stats} further summarizes several statistics of GeoEvidence-6K, including the question vocabulary, token-length distribution, and task distribution.

\begin{figure*}[!t]
    \centering
    \includegraphics[width=\textwidth]{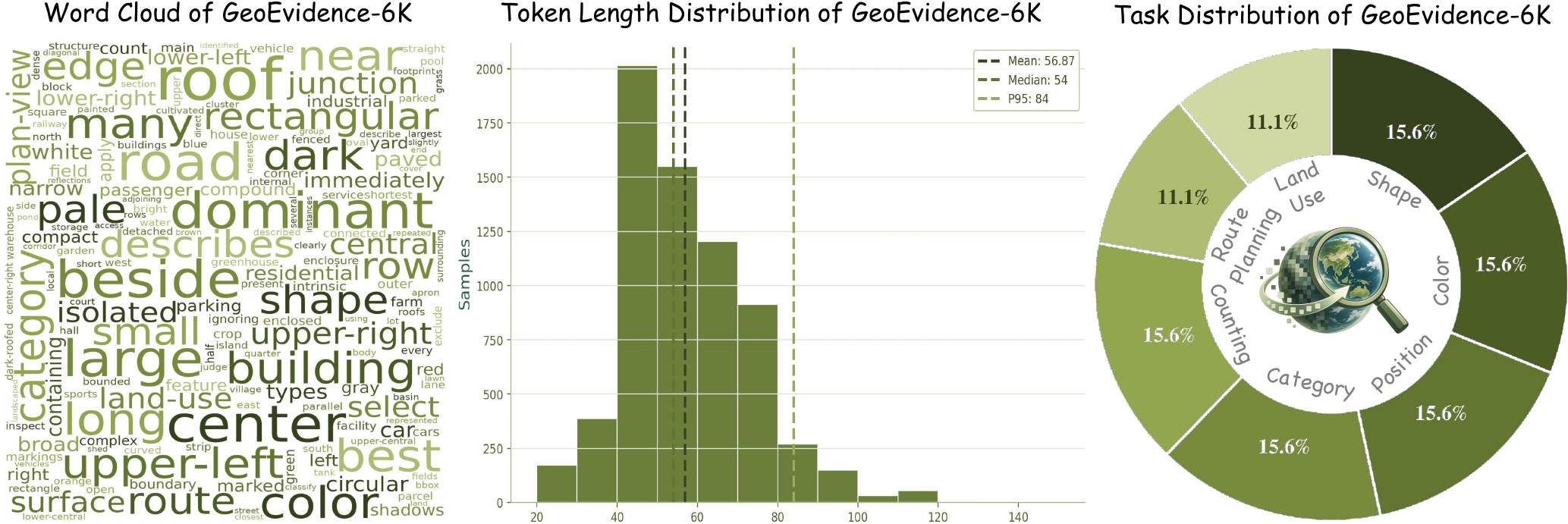}
    \caption{
    Dataset statistics of GeoEvidence-6K.
    Left: word cloud of the question texts.
    Middle: token-length distribution of the questions, with the mean, median, and 95th percentile marked by dashed lines.
    Right: distribution of task categories.
    }
    \label{fig:dataset-stats}
\end{figure*}

As shown in Figure~\ref{fig:dataset-stats}(left), the most frequent words include terms such as \textit{roof}, \textit{parcel}, \textit{road}, \textit{color}, \textit{shape}, \textit{center}, and relative-location expressions such as \textit{upper-right}, \textit{beside}, and \textit{near}. This indicates that GeoEvidence-6K covers diverse UHR remote sensing VQA patterns, including attribute recognition, object counting, spatial localization, and fine-grained geometric reasoning.

Figure~\ref{fig:dataset-stats}(middle) shows that the token-length distribution is concentrated in a moderate range, with a mean of 56.87, a median of 54, and a 95th percentile of 84. Most samples therefore remain relatively compact, while a smaller fraction of longer questions reflects the presence of more compositional and spatially specific instructions.

Figure~\ref{fig:dataset-stats}(right) shows that the task categories are overall well balanced. Five local-evidence tasks---\textit{counting}, \textit{position}, \textit{category}, \textit{color}, and \textit{shape}---each account for about 15.6\% of the dataset, while the two global-context tasks---\textit{route planning} and \textit{land use}---each account for about 11.1\%. This composition provides balanced supervision across both local fine-grained perception and global scene-level understanding.
\section{Implementation Details}
\label{sec:implementation-details}

\begin{figure*}[h]
    \centering
    \includegraphics[width=\textwidth]{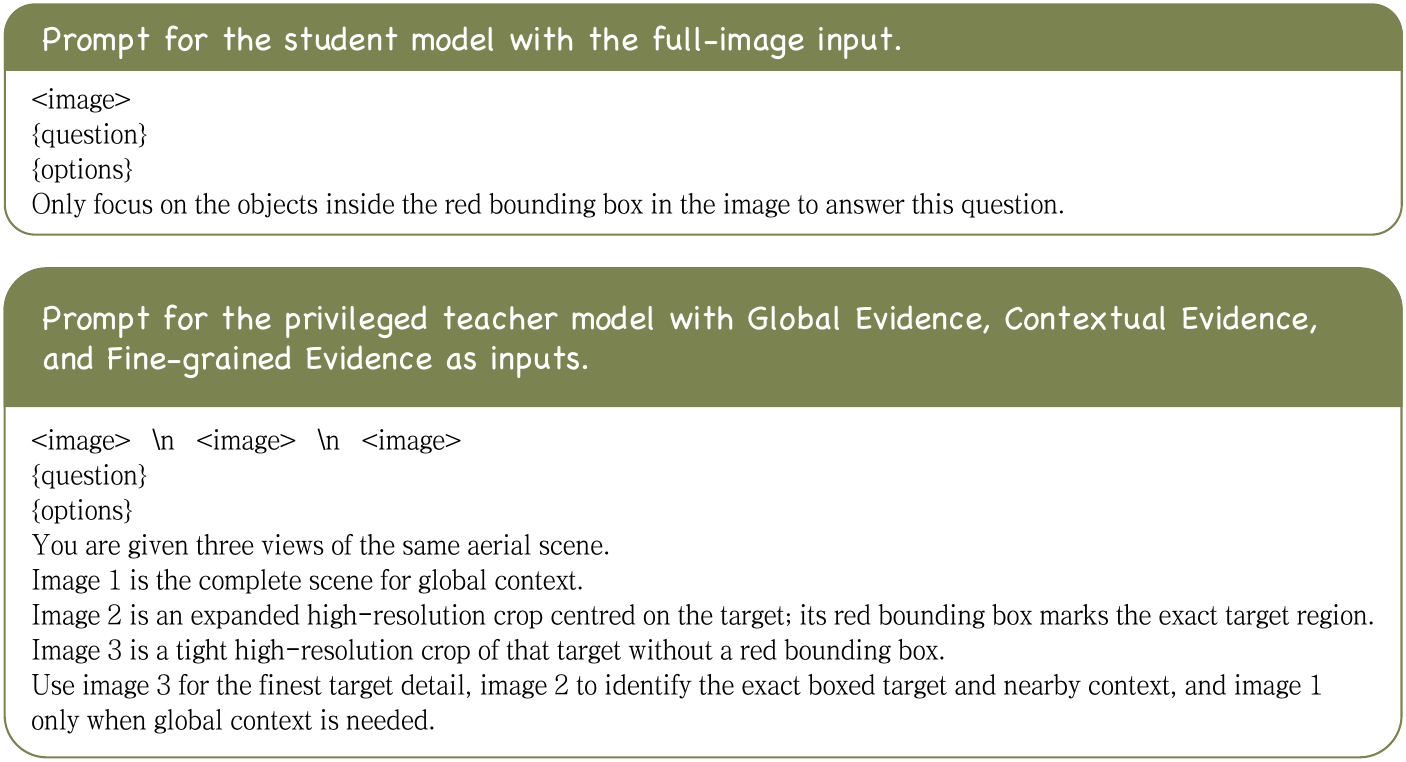}
    \caption{
    Prompt templates for the student and privileged teacher in
    \textcolor{armygreen}{RS-OPSD}.
    }
    \label{fig:training-prompts}
\end{figure*}

\paragraph{Training framework.}
We implement \textcolor{armygreen}{RS-OPSD} using the \texttt{verl}~\citep{verl} training framework. For each training sample, the student performs a single on-policy rollout ($n=1$), which is then reused for privileged-teacher evaluation, CAD computation, and reference-model regularization.

\paragraph{Visual preprocessing.}
During training, all visual inputs are resized while preserving the original aspect ratio, with the longest image side capped at 2,048 pixels. The same preprocessing strategy is applied to the Global Evidence, Contextual Evidence, and Fine-grained Evidence. For training samples with localized evidence annotations, the student Global Evidence contains a rendered red bounding box indicating the annotated evidence region, while the privileged teacher additionally receives the corresponding contextual and fine-grained views. At inference time, we do not introduce any additional bounding boxes, crop coordinates, or localization cues beyond those originally provided by the benchmark. Specifically, MME-RealWorld-RS and LRS-VQA are evaluated using their original unmarked images, whereas a subset of XLRS-Bench samples natively contains red-circle annotations in the released benchmark images.

\paragraph{Training prompts.}
To ensure reproducibility, Figure~\ref{fig:training-prompts} summarizes the prompt templates used during \textcolor{armygreen}{RS-OPSD} training. The student receives a single full-image input together with the original question, options, and bounding-box instruction when applicable. In contrast, the privileged teacher receives three complementary views---Global Evidence, Contextual Evidence, and Fine-grained Evidence---with an explicit instruction describing the role of each view. The question and answer options are otherwise kept consistent between the student and teacher prompts. All prompts are wrapped with the Qwen chat template before being passed to the model.

\paragraph{Inference efficiency.}
Inference latency is measured on a single NVIDIA H100 GPU with a batch size of 1. We report the average per-sample latency under the same evaluation protocol used for the three UHR remote sensing VQA benchmarks. Since \textcolor{armygreen}{RS-OPSD} introduces no architectural modifications to the underlying vision-language model and requires neither additional visual search nor external tool calls at inference time, the trained model remains directly compatible with standard inference acceleration frameworks such as vLLM~\citep{vllm}, without requiring specialized serving infrastructure. 
\section{Detailed Quantitative Comparison Results}
\label{sec:detailed}

Tables~\ref{tab:xlrs_detailed} and~\ref{tab:detailed_results} provide
fine-grained quantitative comparison results across the three UHR remote sensing VQA benchmarks.

\textbf{XLRS-Bench~\citep{xlrs}.}
XLRS-Bench evaluates eight perception categories---Overall Counting (OC),
Regional Counting (RC), Overall Land Use Classification (OLUC), Regional
Land Use Classification (RLUC), Object Classification (OCC), Object Color
(OCL), Object Motion State (OMS), and Object Spatial Relationship (OSR)---and
five reasoning categories, including Anomaly Detection (AD), Environmental
Conditional Reasoning (ECR), Route Planning (RP), Regional Counting with
Change Detection (RCCD), and Counting with Complex Reasoning (CCR).
As shown in Table~\ref{tab:xlrs_detailed}, \textcolor{armygreen}{RS-OPSD}
achieves the best overall score of 53.1, exceeding the previous best
GeoLLaVA-8K~\citep{geollava8k} by 1.6 points. It ranks first on RC, RLUC, OCC, ECR, and RCCD,
while obtaining the second-best results on OCL and OMS. The improvements
span both perception and reasoning tasks, indicating that the gains are not
concentrated in a single capability.

\textbf{MME-RealWorld-RS~\citep{mmerealword} and LRS-VQA~\citep{lrsvqa}.}
As shown in Table~\ref{tab:detailed_results}, \textcolor{armygreen}{RS-OPSD}
achieves 61.5 on MME-RealWorld-RS, outperforming the previous best
ZoomSearch~\citep{ZoomSearch} by 3.9 points. In particular, it achieves the best results on
Position (76.8) and Color (70.4), while remaining competitive on Count.
On LRS-VQA, \textcolor{armygreen}{RS-OPSD} reaches the best average score of 33.3, improving
over WeaveEarth~\citep{weaveearth} by 2.0 points. Although it does not dominate every individual
subset, it achieves consistently strong results across FAIR, Bridge, and STAR,
leading to the highest overall average.

\begin{table*}[!t]
\centering
\caption{Detailed quantitative results on XLRS-Bench across perception and reasoning sub-tasks. \textcolor{darkgreen}{Green}, \textcolor{darkyellow}{yellow}, and \textcolor{darkblue}{blue} denote the first-, second-, and third-best results, respectively.}
\label{tab:xlrs_detailed}

\setlength{\tabcolsep}{4.0pt}
\renewcommand{\arraystretch}{1.10}

\resizebox{\textwidth}{!}{
\begin{tabular}{lcccccccccccccc}
\toprule

\textbf{Method}
& \multicolumn{8}{c}{\textbf{Perception}}
& \multicolumn{5}{c}{\textbf{Reasoning}}
& \textbf{Avg.} \\

\cmidrule(lr){2-9}
\cmidrule(lr){10-14}

\textbf{Sub-tasks}
& \textbf{OC}
& \textbf{RC}
& \textbf{OLUC}
& \textbf{RLUC}
& \textbf{OCC}
& \textbf{OCL}
& \textbf{OMS}
& \textbf{OSR}
& \textbf{AD}
& \textbf{ECR}
& \textbf{RP}
& \textbf{RCCD}
& \textbf{CCR}
& \\

\midrule

\multicolumn{15}{c}{\textit{Closed-Source Vision Language Models}} \\
\midrule

GPT-4o
& 25.0 & 32.0 & 15.0 & 66.0 & 9.5 & 11.3 & 11.7 & 24.6
& 73.0 & 73.0 & 35.0 & 20.0 & 25.0 & 32.4 \\

Claude 3.7 Sonnet
& 27.6 & 22.7 & 17.4 & 68.4 & 30.5 & 29.9 & 63.6 & 27.6
& 64.8 & 78.4 & 34.5 & 27.8 & 32.6 & 40.5 \\

Gemini 2.5 Pro
& \textcolor{darkgreen}{\textbf{50.0}} & 42.0 & 12.0 & 65.5 & 37.5 & 38.1 & 60.0 & 30.2
& 68.0 & 72.0 & 36.0 & 26.7 & 35.0 & 44.1 \\

\midrule
\multicolumn{15}{c}{\textit{Open-Source Vision Language Models}} \\
\midrule

LLaVA-OV-7B
& 25.0 & 38.0 & 8.0 & 69.5 & 35.9 & 35.3 & 65.0 & 25.2
& \textcolor{darkyellow}{\textbf{76.0}} & \textcolor{darkyellow}{\textbf{83.0}} & 24.0 & 43.3 & 36.0 & 43.4 \\

Qwen2.5-VL-7B
& 33.3 & 40.0 & 31.0 & 77.0 & 40.6 & \textcolor{darkblue}{\textbf{40.5}} & \textcolor{darkblue}{\textbf{66.7}} & \textcolor{darkyellow}{\textbf{36.2}}
& 68.0 & 72.0 & 27.0 & 38.3 & 45.0 & 47.4 \\

Qwen3-VL-8B
& 23.3 & \textcolor{darkblue}{\textbf{45.0}} & 20.0 & \textcolor{darkyellow}{\textbf{82.0}} & \textcolor{darkyellow}{\textbf{45.9}} & \textcolor{darkgreen}{\textbf{44.4}} & \textcolor{darkblue}{\textbf{66.7}} & 30.6
& \textcolor{darkblue}{\textbf{74.0}} & 79.0 & 42.0 & \textcolor{darkblue}{\textbf{48.3}} & \textcolor{darkgreen}{\textbf{55.0}} & \textcolor{darkblue}{\textbf{50.5}} \\

InternVL3-8B
& \textcolor{darkblue}{\textbf{40.0}} & 39.0 & 10.0 & 71.5 & \textcolor{darkblue}{\textbf{44.5}} & 30.8 & 65.0 & 25.2
& \textcolor{darkgreen}{\textbf{77.0}} & \textcolor{darkblue}{\textbf{82.0}} & 36.0 & 21.7 & 50.0 & 45.6 \\

\midrule
\multicolumn{15}{c}{\textit{Remote Sensing Vision Language Models}} \\
\midrule

GeoChat
& 16.7 & 29.0 & 2.0 & 23.0 & 21.1 & 16.8 & 35.0 & 24.2
& 33.0 & 43.0 & 10.0 & -- & 21.0 & 22.9 \\

VHM
& 18.3 & 33.0 & 5.0 & 38.5 & 27.5 & 25.4 & 35.0 & 33.0
& 53.0 & 58.0 & 39.0 & 26.7 & 34.0 & 32.8 \\

\midrule
\multicolumn{15}{c}{\textit{Specialized Methods for UHR Remote Sensing VQA}} \\
\midrule

GeoLLaVA-8K
& 26.7 & 38.0 & \textcolor{darkyellow}{\textbf{49.0}} & 69.0 & 41.6 & 31.6 & 65.0 & \textcolor{darkblue}{\textbf{35.0}}
& 67.0 & 78.0 & \textcolor{darkgreen}{\textbf{66.0}} & \textcolor{darkyellow}{\textbf{50.0}} & \textcolor{darkyellow}{\textbf{52.0}} & \textcolor{darkyellow}{\textbf{51.5}} \\

UHR-BAT
& 21.7 & 33.0 & \textcolor{darkgreen}{\textbf{50.0}} & 55.5 & 43.5 & 33.8 & 65.0 & \textcolor{darkgreen}{\textbf{44.8}}
& 62.0 & 71.0 & \textcolor{darkyellow}{\textbf{54.0}} & 46.7 & \textcolor{darkblue}{\textbf{51.0}} & 48.6 \\

\addlinespace[2pt]
\grayline
\addlinespace[2pt]

ZoomSearch
& \textcolor{darkyellow}{\textbf{46.7}} & 42.0 & 12.0 & 70.5 & \textcolor{darkblue}{\textbf{44.5}} & 35.8 & \textcolor{darkgreen}{\textbf{75.0}} & 32.2
& \textcolor{darkblue}{\textbf{74.0}} & \textcolor{darkblue}{\textbf{82.0}} & 29.0 & 43.3 & 37.0 & 48.0 \\

WeaveEarth
& 20.0 & \textcolor{darkyellow}{\textbf{47.0}} & 24.0 & \textcolor{darkblue}{\textbf{79.0}} & 43.1 & 38.0 & \textcolor{darkblue}{\textbf{66.7}} & 26.8
& 69.0 & 81.0 & \textcolor{darkblue}{\textbf{46.0}} & \textcolor{darkblue}{\textbf{48.3}} & \textcolor{darkgreen}{\textbf{55.0}} & 49.5 \\

\addlinespace[2pt]
\grayline
\addlinespace[2pt]

ZoomEarth
& 18.3 & 35.0 & 30.0 & 61.5 & 39.2 & 32.4 & 31.7 & 20.8
& 70.0 & 75.0 & 38.0 & 28.3 & 30.0 & 39.2 \\

\midrule
\multicolumn{15}{c}{\textit{Ours}} \\
\midrule

\textcolor{armygreen}{RS-OPD-\textit{Lite}}
& 25.0 & \textcolor{darkblue}{\textbf{45.0}} & 30.0 & 72.0 & 31.1 & 36.8 & 63.3 & 28.0
& \textcolor{darkblue}{\textbf{74.0}} & \textcolor{darkyellow}{\textbf{83.0}} & 20.0 & 36.7 & 50.0 & 45.8 \\

\textcolor{armygreen}{RS-OPSD}
& 28.3 & \textcolor{darkgreen}{\textbf{55.0}} & \textcolor{darkblue}{\textbf{33.0}} & \textcolor{darkgreen}{\textbf{85.0}} & \textcolor{darkgreen}{\textbf{46.8}} & \textcolor{darkyellow}{\textbf{41.9}} & \textcolor{darkyellow}{\textbf{71.7}} & 29.2
& \textcolor{darkblue}{\textbf{74.0}} & \textcolor{darkgreen}{\textbf{84.0}} & 45.0 & \textcolor{darkgreen}{\textbf{51.7}} & 45.0 & \textcolor{darkgreen}{\textbf{53.1}} \\

\bottomrule
\end{tabular}
}
\end{table*}

\begin{table*}[!t]
\centering
\caption{
Detailed quantitative results on MME-RealWorld-RS, LRS-VQA, and XLRS-Bench.
Average inference speed is measured in s/sample.
}
\label{tab:detailed_results}

\setlength{\tabcolsep}{4.0pt}
\renewcommand{\arraystretch}{1.10}

\resizebox{\textwidth}{!}{
\begin{tabular}{l cccc cccc c cc}
\toprule

\textbf{Method}
& \multicolumn{4}{c}{\textbf{MME-RealWorld-RS}}
& \multicolumn{4}{c}{\textbf{LRS-VQA}}
& \multicolumn{1}{c}{\textbf{XLRS-Bench}}
& \multicolumn{2}{c}{\textbf{Total}}
\\

\cmidrule(lr){2-5}
\cmidrule(lr){6-9}
\cmidrule(lr){10-10}
\cmidrule(lr){11-12}

\textbf{Sub-set}
& \textbf{Position}
& \textbf{Color}
& \textbf{Count}
& \textbf{Avg.}
& \textbf{FAIR}
& \textbf{Bridge}
& \textbf{STAR}
& \textbf{Avg.}
& \textbf{Avg.}
& \textbf{Accuracy}
& \textbf{Speed}
\\

\midrule
\multicolumn{12}{c}{\textit{Closed-Source Vision Language Models}} \\
\midrule

GPT-4o
& 36.4 & 32.4 & 15.9 & 28.2
& 22.2 & 31.8 & 27.4 & 27.1
& 32.4 & 29.2 & -- \\

Claude 3.7 Sonnet
& 56.6 & 52.1 & 28.5 & 45.7
& 25.7 & 29.5 & 30.1 & 28.4
& 40.5 & 38.2 & -- \\

Gemini 2.5 Pro
& 60.6 & 60.3 & 34.3 & 51.7
& \textcolor{darkblue}{\textbf{28.9}} & 31.2 & 32.7 & \textcolor{darkblue}{\textbf{30.9}}
& 44.1 & 42.2 & -- \\

\midrule
\multicolumn{12}{c}{\textit{Open-Source Vision Language Models}} \\
\midrule

LLaVA-OV-7B
& 64.8 & 61.4 & 33.8 & 53.3
& 20.6
& \textcolor{darkyellow}{\textbf{35.1}}
& 26.1 & 27.3
& 43.4 & 41.3 & 2.38 \\

Qwen2.5-VL-7B
& 55.9 & 48.5 & 18.6 & 41.0
& 23.9
& 34.2
& 30.2 & 29.4
& 47.4 & 39.3 & 1.87 \\

Qwen3-VL-8B
& 56.4
& 48.1
& 21.3
& 41.9
& 24.6
& \textcolor{darkgreen}{\textbf{38.2}}
& 27.5
& 30.1
& \textcolor{darkblue}{\textbf{50.5}}
& 40.8
& 1.75 \\

InternVL3-8B
& \textcolor{darkblue}{\textbf{71.0}} & \textcolor{darkblue}{\textbf{62.6}} & 31.4 & 55.0
& 24.8 & 33.4 & 29.3 & 29.2
& 45.6 & 43.3 & \textcolor{darkyellow}{\textbf{1.25}} \\

\midrule
\multicolumn{12}{c}{\textit{Remote Sensing Vision Language Models}} \\
\midrule

GeoChat
& 25.1 & 23.1 & 15.7 & 21.3
& 20.2 & 24.5 & 13.8 & 19.5
& 22.9 & 21.2 & 1.33 \\

VHM
& 35.2 & 20.3 & 16.8 & 24.1
& 24.3 & 27.5 & 28.3 & 26.7
& 32.8 & 27.9 & 2.64 \\

\midrule
\multicolumn{12}{c}{\textit{Specialized Methods for UHR Remote Sensing VQA}} \\
\midrule

GeoLLaVA-8K
& 34.9 & 27.9 & 22.3 & 28.4
& 21.8 & 29.2 & 26.4 & 25.8
& \textcolor{darkyellow}{\textbf{51.5}}
& 35.2 & \textcolor{darkblue}{\textbf{1.30}} \\

UHR-BAT
& 44.0 & 42.0 & 14.0 & 33.3
& 16.6 & 23.5 & 20.0 & 20.0
& 48.6 & 34.0 & 7.40 \\

\addlinespace[2pt]
\grayline
\addlinespace[2pt]

ZoomSearch
& 67.6
& \textcolor{darkyellow}{\textbf{66.1}}
& \textcolor{darkyellow}{\textbf{39.2}}
& \textcolor{darkyellow}{\textbf{57.6}}
& 25.9 & 31.2 & 33.5 & 30.2
& 48.0
& \textcolor{darkyellow}{\textbf{45.3}}
& 43.5 \\

WeaveEarth
& 51.4 & 27.7
& \textcolor{darkgreen}{\textbf{62.6}}
& 47.2
& \textcolor{darkgreen}{\textbf{31.0}}
& 26.1
& \textcolor{darkgreen}{\textbf{36.7}}
& \textcolor{darkyellow}{\textbf{31.3}}
& 49.5 & 42.7 & 2.94 \\

\addlinespace[2pt]
\grayline
\addlinespace[2pt]

ZoomEarth
& 46.6 & 41.9 & 27.6 & 38.7
& 20.0 & 23.2 & 21.6 & 21.6
& 39.2 & 33.2 & 14.2 \\

\midrule
\multicolumn{12}{c}{\textit{Ours}} \\
\midrule

\textcolor{armygreen}{RS-OPD-\textit{Lite}}
& \textcolor{darkyellow}{\textbf{74.2}}
& 61.9
& 32.5
& \textcolor{darkblue}{\textbf{56.2}}
& 27.9 & 29.6 & \textcolor{darkblue}{\textbf{33.9}} & 30.5
& 45.8
& \textcolor{darkblue}{\textbf{44.2}}
& \textcolor{darkgreen}{\textbf{1.09}} \\

\textcolor{armygreen}{RS-OPSD}
& \textcolor{darkgreen}{\textbf{76.8}}
& \textcolor{darkgreen}{\textbf{70.4}}
& \textcolor{darkblue}{\textbf{37.4}}
& \textcolor{darkgreen}{\textbf{61.5}}
& \textcolor{darkyellow}{\textbf{29.4}}
& \textcolor{darkblue}{\textbf{34.9}}
& \textcolor{darkyellow}{\textbf{35.6}}
& \textcolor{darkgreen}{\textbf{33.3}}
& \textcolor{darkgreen}{\textbf{53.1}}
& \textcolor{darkgreen}{\textbf{49.3}}
& 1.58 \\

\bottomrule
\end{tabular}
}
\end{table*}

\textbf{Overall accuracy and efficiency.}
Averaged across the three benchmarks, \textcolor{armygreen}{RS-OPSD} reaches 49.3, a 4.0-point improvement over the strongest competing method.
Meanwhile, \textcolor{armygreen}{RS-OPD-\textit{Lite}} achieves an average
accuracy of 44.2 with only a 2B student, surpassing all evaluated
8B-scale baselines. It also achieves the lowest average inference time at 1.09 s/sample, compared with 1.25 s/sample for the second-fastest
method. The full \textcolor{armygreen}{RS-OPSD} remains efficient at 1.58 s/sample, even
faster than its Qwen3-VL-8B~\citep{qwen3vl} base model at 1.75 s/sample. These results further
show that internalizing privileged visual information through OPSD provides a
favorable balance between accuracy and inference efficiency.
\section{Limitations and Future Work}
\label{sec:limitations}

The current formulation of \textcolor{armygreen}{RS-OPSD} relies on explicit supervision at two stages. First, target-region annotations are required to construct the privileged visual inputs used by CPVP. Second, ground-truth answers are used in CAD to assess teacher reliability and determine the correctness of student rollouts. Although these annotations enable reliable privileged distillation, acquiring them for UHR remote sensing imagery introduces substantial annotation cost and limits the scalability of the framework.

A natural extension is to explore more self-supervised forms of OPSD that reduce or remove these dependencies. Future work may investigate automatic discovery of question-relevant visual privilege without target-region annotations, as well as teacher reliability estimation without ground-truth answer constraints. More broadly, combining model-intrinsic evidence discovery with self-verification could enable privileged self-distillation to scale to larger unlabeled UHR datasets and broader high-resolution vision-language tasks.

Our \textcolor{armygreen}{RS-OPD-\textit{Lite}} experiments further suggest that visual privilege remains beneficial beyond the self-distillation setting, where a larger privileged teacher transfers knowledge to a smaller student. This observation opens another direction for \emph{privilege-enhanced model distillation}, where additional training-time information can improve knowledge transfer across model scales while retaining a lightweight student at inference time. Exploring such privileged distillation may provide a promising route toward more accurate and efficient vision-language models.

\end{document}